\documentclass[10pt,twocolumn,letterpaper]{article}

\usepackage[pagenumbers]{cvpr} % To force page numbers, e.g. for an arXiv version

\usepackage{graphicx}
\usepackage{amsmath}
\usepackage{amssymb}
\usepackage{booktabs}
\usepackage{lipsum}
\usepackage{multirow}
\usepackage{makecell}
\usepackage[accsupp]{axessibility}

\usepackage{ctable}
\usepackage{siunitx}
\usepackage{enumitem} %for leftmargin of enumerate

\usepackage[pagebackref,breaklinks,colorlinks]{hyperref}

\usepackage[capitalize]{cleveref}
\crefname{section}{Sec.}{Secs.}
\Crefname{section}{Section}{Sections}
\Crefname{table}{Table}{Tables}
\crefname{table}{Tab.}{Tabs.}

\def\cvprPaperID{10875} % *** Enter the CVPR Paper ID here
\def\confName{CVPR}
\def\confYear{2022}

\begin{document}

\title{\emph{MMG-Ego4D}: \underline{M}ulti-\underline{M}odal \underline{G}eneralization in Egocentric Action Recognition}

%%%%% Title Suggestions %%%%%%%%%
% Navigating the world with broken sensors: Improving generalization ability of multimodal systems. 

% \author{Xinyu Gong$^*$ 
% \and Sreyas Mohan$^*$
% \and Naina Dhingra
% \and Jean-Charles Bazin
% \and Yilei Li
% \and Zhangyang Wang
% \and Rakesh Ranjan  
% }
\author{Xinyu Gong$^2$\thanks{Equal contribution} \thanks{Work done during an internship at Meta Reality Labs.} , Sreyas Mohan$^1$$^*$ , Naina Dhingra$^1$, Jean-Charles Bazin$^1$, Yilei Li$^1$, \\Zhangyang Wang$^2$, Rakesh Ranjan$^1$  \\
$^1$Meta Reality Labs, $^2$The University of Texas at Austin 
}

% % omit the following lines up until the closing "}".
% % Additional authors and addresses can be added with "\and",
% \and
\maketitle

%%%%%%%%% ABSTRACT
\begin{abstract}
In this paper, we study a novel problem in egocentric action recognition, which we term as \textit{``Multimodal Generalization''} (\textbf{MMG}). MMG aims to study how systems can generalize when data from certain modalities is limited or even completely missing. We thoroughly investigate MMG in the context of standard supervised action recognition and the more challenging few-shot setting for learning new action categories. MMG consists of two novel scenarios, designed to support security, and efficiency considerations in real-world applications: (1) missing modality generalization where some modalities that were present during the train time are  missing during the inference time, and (2) cross-modal zero-shot generalization, where the modalities present during the inference time and the training time are \textit{disjoint}. To enable this investigation, we construct a new dataset \textit{MMG-Ego4D} containing data points with video, audio, and inertial motion sensor (IMU) modalities. Our dataset is derived from \textit{Ego4D}~\cite{ego4d} dataset, but processed and thoroughly re-annotated by human experts to facilitate research in the MMG problem. We evaluate a diverse array of models on \textit{MMG-Ego4D} and propose new methods with improved generalization ability. In particular, we introduce a new fusion module with modality dropout training, contrastive-based alignment training, and a novel cross-modal prototypical loss for better few-shot performance. We hope this study will serve as a benchmark and guide future research in multimodal generalization problems. The benchmark and code are available at \href{https://github.com/facebookresearch/MMG_Ego4D}{https://github.com/facebookresearch/MMG\_Ego4D}
\end{abstract}

%%%%%%%%% BODY TEXT
\section{Introduction}
\label{sec:intro}

Action recognition systems are typically trained on data captured from a third-person or spectator perspective~\cite{kinetics, poppe2010survey}. However, in areas such as robotics and augmented reality, we capture data through the eyes of agents, \textit{i.e.}, in a first-person or egocentric perspective. With head-mounted devices such Ray-Ban Stories becoming popular, action recognition from egocentric videos is critical to enable downstream applications, such as contextual recommendations or reminders. However, egocentric action recognition is fundamentally different and more challenging~\cite{damen2018scaling, damen2022rescaling, li2018eye, pirsiavash2012detecting}. While third-person video clips are often curated, egocentric video clips are uncurated and have low-level corruptions, such as large motion blur due to head motion. Moreover, egocentric perception requires a careful understanding of the camera wearer's physical surroundings, and must interpret the objects and interactions from the wearer's perspective. 

%%%%%%%%%%% multimodal demonstration %%%%%%%%%%
\begin{figure}[!t]
    \centering
    \includegraphics[width=1.0\linewidth]{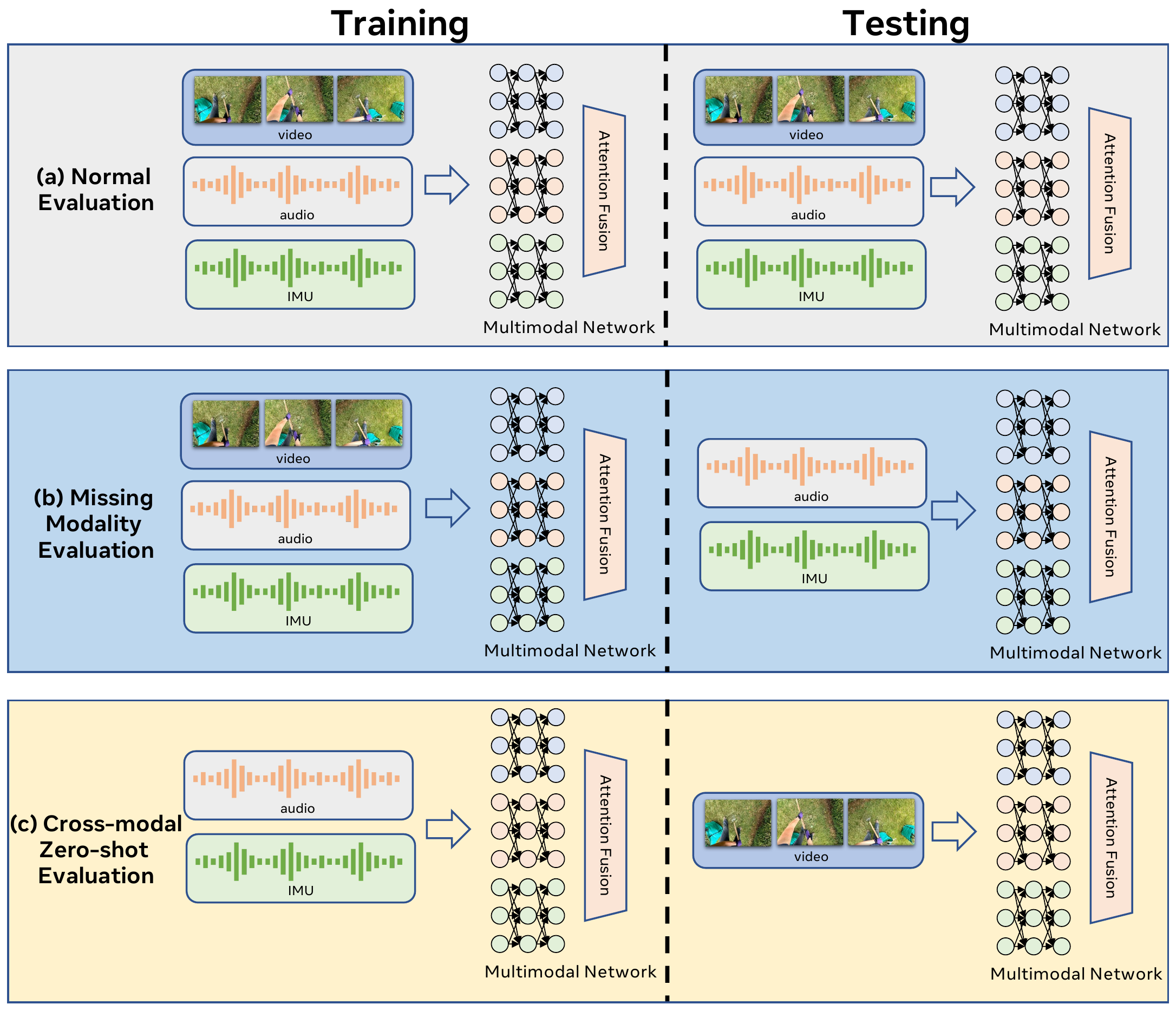}
       \caption{\textbf{Overview of \emph{MMG-Ego4D} challenge}. In a typical evaluation setting (a) networks are trained for the supervised setting or the few-shot setting using training/support sets with data from all modalities and evaluated on data points with all modalities. However, there can often be a mismatch between training and testing modalities. Our proposed challenge contains two tasks to mimic these settings. In (b) \emph{missing modality evaluation}, the model can only use a subset of training modalities for inference. In (c) \emph{ Cross-modal zero-shot evaluation}, the models are on modalities unseen during training.}
    \label{fig:teaser}
\end{figure}

%%%%%%%%%%% multimodal demonstration %%%%%%%%%%
\begin{figure}[!t]
    \centering
    \includegraphics[width=1.0\linewidth]{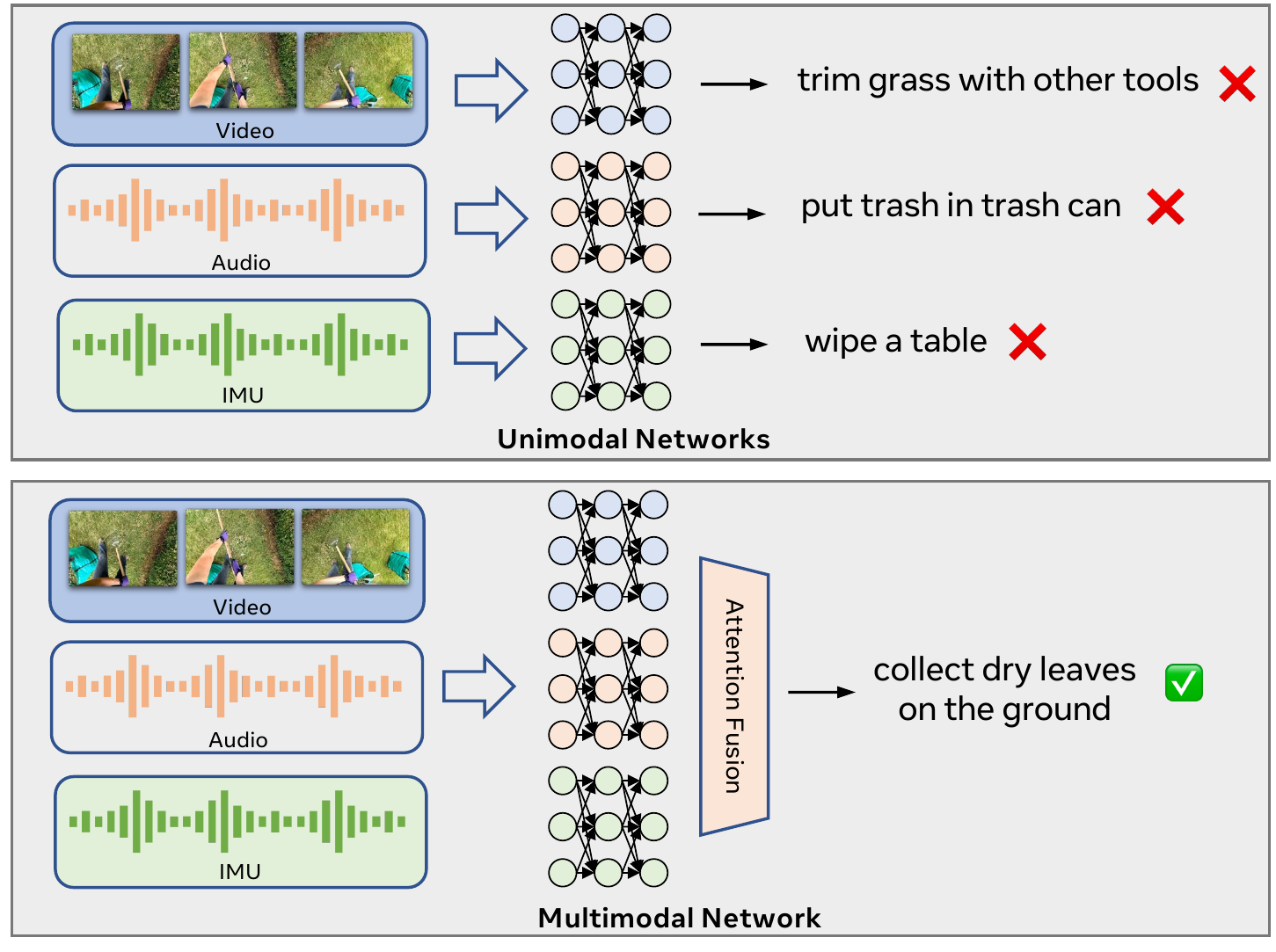}
       \caption{\textbf{Multimodal data is crucial for egocentric perception}. Input data consists of three modalities: video, audio, and IMU. (top) Video action recognition identifies the clip with a tool and much grass in the background as the class \emph{trim grass with other tools}. Audio action recognition system classifies the periodic rubbing sound as \emph{put trash in a trash can}. The IMU model classifies the head movement action into the class \emph{wipe a table}.
       (bottom) Multimodal action recognition system correctly combines the video feed and audio feed and identifies the activity as \emph{collect dry leaves on the ground}. }
    \label{fig:multimodal}
\end{figure}
%%%%%%%%%%%%%%%%%%

Recognizing egocentric activity exclusively from one modality can often be ambiguous. This is because we want to perceive what the device's wearer is performing instead of what the camera feed is capturing. To this end, Multimodal information can be crucial for understanding and disambiguating the user's intent or action. We demonstrate it through an example in Fig.~\ref{fig:multimodal}. In the example, the video feed shows a tool in the background of the grassland.  An activity recognition model exclusively based on video recognizes it as the class \emph{trim grass with other tools}. Similarly,  a model exclusively trained in audio identifies the rubbing sounds in the clip as the class \emph{put trash in a trash can}, and an IMU model mistakes the head motion as \emph{wipe a table}. However, a multimodal system correctly identifies the class as \emph{collect dry leaves on the ground} by combining video, audio, and IMU signals. 

While using multimodal information is essential to achieve state-of-the-performance, it also presents a unique challenge - \emph{we may not be able to use all modalities in the real world} due to security or efficiency considerations. For example, a user might be located in a sensitive environment and decide to turn off the camera due to security concerns. Similarly, users may turn off microphones so that their voices are not heard. In these situations, multimodal systems must be able \emph{generalize to missing modalities} (Fig. \ref{fig:teaser} (b)), \textit{i.e.}, work with an incomplete set of modalities at inference, and make a robust prediction. These challenges are not just limited to inference time but could manifest in restrictions during training. For example, if a user has to train a system, often in a few-shot setting, computationally expensive modalities like video are best trained on the cloud. However, the user might prefer that their data stays on the device. However, the video will consume $60\times$ more storage, and $43\times$ more compute compared to cheaper modalities like IMU (see Tab.~\ref{tab:data_size}), significantly increasing the difficulty of training on devices with limited compute and storage. 
In this situation, we may want to enable training with computationally less demanding modalities like audio while maintaining the flexibility of performing inference on more informative modalities like video. 
Multimodal systems should \emph{robustly generalize across modalities}. 

\begin{table}[!t]
\begin{center}
\resizebox{0.85\linewidth}{!}{
\begin{tabular}{c|ccc} \toprule
\bf Modality & video & audio & IMU \\
\midrule
\bf Memory per second of data (KB) & $593.92$ & $62.76$ & $9.44$ \\ 
\bf Typical model FLOPs (G) & $70.50$ & $42.08$ & $1.65$ \\ 
\bottomrule
\end{tabular}
}
\caption{\textbf{Compute and memory cost for different modalities}. Memory used per second for each modality is computed by averaging the memory used by $1000$ data points drawn randomly from Ego4D~\cite{ego4d}. The provided compute number corresponds to the forward pass cost of MViT~\cite{mvit} for video, AST~\cite{ast} for audio, and a ViT~\cite{vit} based transformer model for IMU data. 
}
\label{tab:data_size}
\end{center}
\vspace{-3pt}
\end{table}
\vspace{-2pt}
In this work, we propose \emph{MMG-Ego4D}: a challenge designed to measure the generalization ability of egocentric activity recognition models. Our challenge consists of two novel tasks: (1) \emph{missing modality generalization} aimed at measuring the generalization ability of models when evaluated on an incomplete set of modalities  (shown in Fig. \ref{fig:teaser} (b)), and (2) \emph{cross-modal zero-shot  generalization} aimed at measuring the generalization ability of models in generalizing to unseen modalities during test time (shown in Fig. \ref{fig:teaser} (c)). We evaluate several widely-used architectures using this benchmark and introduce a novel approach that enhances generalization capability in the \emph{MMG-Ego4D} challenge, while also improving performance in standard full-modalities settings. Our primary contributions are:
\begin{itemize}[leftmargin=*,topsep=0pt]
    \item \textbf{MMG Problem}. We present \emph{MMG}, a novel and practical problem with two tasks, \emph{missing modality generalization} and \emph{cross-modal zero-shot  generalization}, for evaluating the generalization ability of multimodal action recognition models. These tasks are designed to support real-world security and efficiency considerations, and we define them in both supervised and more challenging few-shot settings.
    
    \item \textbf{MMG-Ego4D Dataset}. To facilitate the study of MMG problem in ego-centric action recognition task, we introduce a new dataset, \emph{MMG-Ego4d}, which is derived from Ego4D~\cite{ego4d} dataset by preprocessing the data points and thoroughly re-annotating by \emph{human experts} to suit the task. To the best of our knowledge, this is the first work to introduce these novel evaluation tasks and a benchmark challenge of its kind.

    \item \textbf{Strong Baselines}. We present a new method that achieves strong performance on the generalization ability benchmark and also improves the performance under the normal full-modalities setting. Our method employs a Transformer-based fusion module, which allows for flexible input of different modalities. We employ a cross-modal contrastive alignment loss to project features of different modalities into a unified space.
    Finally, a novel loss function is introduced, which is called \textit{cross-modal prototypical loss}, achieving state-of-the-art results in multimodal few-shot settings. Extensive ablation studies are performed to identify each proposed component's contribution.
\end{itemize}

\section{Related Work}

\noindent \textbf{Multimodal egocentric action recognition}. Action recognition systems are typically trained on video~\cite{feichtenhofer2017spatiotemporal, feichtenhofer2016convolutional, x3d, slowfast, gong2021searching, tran2018closer,wang2018appearance, mvit}. However, for \emph{egocentric} activity recognition (\textit{i.e.}, first-person perspective, or recognizing the activity the user wearing the capturing device is performing), complimentary multimodal information is essential for identifying the correct activity (see Fig.~\ref{fig:multimodal}). Previous methods for multimodal fusion in egocentric activity recognition have ranged from simple concatenation~\cite{poria2016convolutional,xiao2020audiovisual} to tensor decomposition~\cite{liu2018efficient}, with some recent studies adopting transformer-based architectures~\cite{akbari2021vatt,missingmodality,nagrani2021attention,kim2021vilt} that have shown promising results.
In this work, we utilize a Transformer-based fusion module and a modality dropout training strategy to further improve performance on \emph{MMG} tasks.

\noindent \textbf{Generalizability of multimodal models}. 
As one of the tasks belonging to \emph{MMG} problem, the missing modality problem has been studied by a few work recently~\cite{missingmodality,tsai2018learning, pahde2019self, ma2021smil,lee2019audio,tran2017missing}. However, most work focuses on the bimodal situation. \cite{tsai2018learning} solve the multimodal image (two domains) classification problem by learning factorized multimodal representations. \cite{ma2021smil} addresses the audio-visual classification problem leveraging a Bayesian meta-learning framework.
\cite{missingmodality} specifically investigate the robustness of the multimodal transformer model to missing-modality data on the text-visual classification task, and improve robustness via multi-task learning and a searched optimal fusion strategy. Cross-modal zero-shot action recognition is still an under-explored new problem. It is related to the cross-modal retrieval problem~\cite{zhen2019deep,thomas2022emphasizing,zeng2022x,xuan2023dissecting}, while the latter focuses on how to measure the feature similarity across different modalities.

\noindent \textbf{Multimodal few-shot learning}. Multimodal few-shot learning~\cite{pahde2019self,tsimpoukelli2021multimodal,ma2022multimodality,pahde2018cross,eloff2019multimodal} is an emerging research area that aims to enable machine learning models to recognize and classify new objects based on limited examples from multiple modalities. Existing research in few-shot learning has predominantly focused on a single modality, like image~\cite{nichol2018first,Chen_2019_CVPR,Gidaris_2019_CVPR,Sun_2019_CVPR,Tokmakov_2019_ICCV,Li_2019_ICCV,Ravichandran_2019_ICCV,Dvornik_2019_ICCV,Gidaris_2019_ICCV,das2021importance,zhu2021few} or language~\cite{brown2020language,zhao2021calibrate,wang2021entailment,gu2021ppt,winata2021language}. However, there has been an increasing interest in extending few-shot learning to multimodal scenarios. Pioneering work in this area includes using text-conditional GANs to augment data via hallucinating images, as demonstrated in \cite{pahde2018cross,pahde2019self}.  
Eloff \textit{et al.}~\cite{eloff2019multimodal} utilize a siamese network for a one-shot cross-modal matching problem on speech and image modalities. 

\noindent \textbf{Datasets and benchmarks}. Availability of datasets and clearly defined benchmark tasks have been a driving factor in improving performance in use cases like classification~\cite{imagenet}, detection~\cite{pascal}, segmentation~\cite{coco} and action recognition~\cite{activitynet, pascal, ava, kinetics}. While performance on these tasks has often surpassed human performance~\cite{he2015delving}, researchers have shown that state-of-the-art methods are often fragile~\cite{hosseini2017google} and do not generalize well to slightly different data points like corruptions~\cite{dodge2017study, geirhos2017comparing} or adversarial examples~\cite{kurakin2016adversarial, carlini2018ground}. Having clearly defined benchmarks, datasets, and tasks to measure the generalization ability has greatly contributed to  driving robustness research~\cite{imagenetc, hirsch2000aurora, hirsch2002experimental}. Our proposed benchmark and dataset, \emph{Ego4D-MMG}, is the first benchmark designed specifically to measure multimodal generalization ability in egocentric action recognition. We hope this benchmark will spur progress in MMG tasks and encourage the development of safety-aware generalizable models.

\vspace{-5pt}
\section{Proposed Benchmark: \emph{MMG-Ego4D}}

\subsection{Overview}
\label{sec:overview}

\noindent \textbf{Preliminaries}. We use the term ``supervised setting" to refer to the regular action recognition task where a large number of labeled training data (training set) are available. During test time, the goal is to classify each testing data point (testing set) into one of the training labels. In contrast, in the few-shot setting, only a few labeled training data are available. The goal at test time is the same as in the supervised setting. In practice, we refer to the training and testing sets in the few-shot setting as the support and query sets, respectively. The collection of support and query sets together is called an ``episode''~\cite{maml,vinyals2016matching,prototypical}. The terms training modalities and testing modalities refer to the available modalities during supervised training and testing, respectively, in the supervised setting. In the few-shot setting, they refer to the support and query modalities.

\noindent \textbf{Data}. The \emph{MMG-Ego4D} dataset comprises data points with three modalities - video, audio, and inertial motion sensors (IMU) sourced from the Ego4D dataset~\cite{ego4d} (we illustrate why we do not choose other datasets in supplementary). The IMU data contains signals obtained from the accelerometer and gyroscope. We use approximately 202 hours of data from the Ego4D dataset to create our benchmark: 167 hours of unlabelled temporal-aligned Video-Audio-IMU data and 35 hours of labeled temporal-aligned data. We perform several steps to make the data suitable for our benchmark.

First, we identify timestamps in the data where the activity occurs and standardize each data point to five seconds, drawing data from the Ego4D Moments track~\cite{ego4d}. IMU and video data are subsampled to 200 Hz and 4 FPS.

Second, several data points in Ego4D contain multiple labels, primarily resulting from (1) multiple activities being performed and (2) activities being related to each other in a hierarchy (e.g., mixing ingredients vs. cooking). We used the WordNet hierarchy as a heuristic to consolidate the label space, using human annotators to scrutinize the labels. If annotators could not conclusively identify a single correct label, we discarded that data point from our benchmark.

Finally, we created the \emph{MMG-Ego4D} few-shot benchmark with two main criteria. Firstly, it must consist of two semantically-disjoint class sets, namely the base classes and novel classes. Secondly, data points from the same original clip cannot be present in both the base and novel classes. We accomplished this in two steps. Initially, the annotators manually split the 79 labels into 65 base classes and 14 novel classes, ensuring that no semantically similar labels were included in the few-shot evaluation benchmark. Then, we confirmed that data points from the same underlying clip were not present in both the base and novel classes. We used the base classes as the training set for the supervised task and drew additional data from the Ego4D Moments track to form the corresponding testing set for the supervised task.

\subsection{Proposed \emph{MMG} Tasks}
The goal of \emph{MMG-Ego4D} is to evaluate the generalization ability of machine learning algorithms in situations where there is a mismatch between training and testing modalities. Humans can deal with missing modalities quite well. For example, we can identify an action from just a video. Similarly, even if a concept was introduced to us only using video, we can often identify this concept using another modality like audio (e.g., a crying baby). In this section, we describe two novel tasks designed to evaluate the generalization ability of multimodal activity recognition systems. These tasks reflect real-world security considerations while using wearable devices. Further, performance on these tasks could also measure how close our current multimodal machine perception is to human perception. The overview of \emph{MMG} tasks is presented in Fig.~\ref{fig:teaser}.

\noindent \textbf{Missing modality evaluation}. This task measures how well models can perform inference using only a \emph{subset} of modalities that were used for training. During inference, maybe due to power or computational constraints, we may only use a subset of the modalities to perform the evaluation. This presents us with variable evaluation settings, and we select some of them to report their results on the benchmark. In the context of the few-shot setting, this task reduces to using a subset of the support modalities as the query modalities.

\noindent \textbf{Zero-Shot Cross-Modal generalization}. This task measures how well models can generalize to unseen modalities. The training and testing modalities are disjoint. In our context, the models may only use IMU and audio for training (training video models is extremely expensive), but they could use video data at test time (the budget for video inference is acceptable and may yield better results). Similarly, in the context of few-shot learning, the support and query modalities are disjoint in this task.

\section{Improving Generalization Performance}
\label{sec:methodology}
This section introduces a strong baseline that achieves high performance on the proposed \emph{MMG-Ego4D} benchmark. Our method comprises three novel components designed to improve the generalization ability of multimodal systems. We begin by presenting an overview of our proposed method, emphasizing the significance of each component, and providing a detailed description of their implementation.

\subsection{Method Overview}
We illustrate the overview of our proposed method pipeline in this section. Under the few-shot setting, all evaluation tasks adopt the same training pipeline, composed of three stages. (1) \textit{unimodal supervised pre-training}: feature extractor for each modality is trained separately. (2) \textit{multimodal supervised pre-training}: a fusion module is attached at the end of unimodal networks to form a multimodal system, which is then trained with a cross-entropy loss and a cross-modal contrastive alignment loss. The latter loss term aims to enhance the multimodal generalizability of the model by constructing a unified feature space for all modalities. (3) \textit{multimodal meta-training}: the multimodal network is meta-trained with prototypical-based loss to further improve the model's cross-modal generalizability. It's worth noting that data of all modalities are used in the above training pipeline of the few-shot setting. The modality restriction of \emph{MMG-Ego4D} tasks is only applied in the support and query set during the few-shot evaluation.

In the supervised setting, the regular and missing modality evaluation settings adopt the same training pipeline, containing (1) \textit{unimodal supervised pre-training} and (2) \textit{multimodal supervised pre-training}, the same as the first two stages of the few-shot setting training pipeline. In contrast, the zero-shot cross-modal setting has a different two-stage training pipeline. (1) \textit{multimodal unsupervised pre-training}: the multimodal network is trained with a cross-modal contrastive alignment loss using unlabeled data, to establish a modality-agnostic unified feature space. (2) \textit{multimodal supervised pre-training}:  the multimodal network is trained using a cross-entropy loss, without the contrastive alignment loss term used in previous settings, as the evaluation modality is absent in the labeled data, due to the restriction of this setting. It is meaningless to construct an alignment between the training modalities. Therefore we choose to build such an alignment using the modality-complete unlabeled data, which also does not violate the rule of this setting. It should be noted that the modality restriction in the \emph{MMG-Ego4D} tasks applies to the labeled training data used in the \textit{multimodal supervised pre-training} stage and the evaluation stage. This is different from the few-shot setting. We have summarized the training pipeline in Tab. \ref{tabl:pipeline}.

%%%%%%%%%%%%%%%%%%%%%%%%%%
\begin{table}
\centering 
\resizebox{1.0\linewidth}{!}{
\begin{tabular}{c|c|cccc}
\toprule
\centering 
    \bf \multirow{3}{*}{Setting} & \bf \multirow{3}{*}{Task} & \bf \multirow{3}{*}{\shortstack[c]{Multimodal \\ unsupervised \\ pre-train}} & \bf \multirow{3}{*}{\shortstack[c]{Unimodal \\ supervised \\ pre-train}} & \bf \multirow{3}{*}{\shortstack[c]{Multimodal \\ supervised \\ train}} & \bf \multirow{3}{*}{\shortstack[c]{Multimodal \\ meta-train}} \\
    & & & & & \\
    & & & & &\\
    \midrule
    
    \multirow{3}{*}{Supervised} & Regular & - & $\mathcal{L}_\text{CE}$ & $\mathcal{L}_\text{CE} + \mathcal{L}_\text{align}$ & - \\ %\cmidrule(lr){2-6}
    & Missing Modal & - & $\mathcal{L}_\text{CE}$ & $\mathcal{L}_\text{CE} + \mathcal{L}_\text{align}$ & -\\  %\cmidrule(lr){2-6}
    & Zero-Shot & $\mathcal{L}_\text{align}$ & - & $\mathcal{L}_\text{CE}$ & -\\
    \midrule 
    
    \multirow{3}{*}{Few-shot} & Regular & - & $\mathcal{L}_\text{CE}$ & $\mathcal{L}_\text{CE} + \mathcal{L}_\text{align}$ & $\mathcal{L}_\text{proto}$\\  %\cmidrule(lr){2-6}
    & Missing Modal & - & $\mathcal{L}_\text{CE}$ & $\mathcal{L}_\text{CE} + \mathcal{L}_\text{align}$ & $\mathcal{L}_\text{proto}$ \\  %\cmidrule(lr){2-6}
    & Zero-Shot & - & $\mathcal{L}_\text{CE}$ & $\mathcal{L}_\text{CE} + \mathcal{L}_\text{align}$ & $\mathcal{L}_\text{proto}$ \\
    \bottomrule
\end{tabular}
}
\caption{\textbf{Training pipelines of supervised \& few-shot settings}. $\mathcal{L}_\text{CE}$ denotes the cross-entropy loss. $\mathcal{L}_\text{align}$ and $\mathcal{L}_\text{proto}$ are cross-modal contrastive alignment loss and cross-modal prototypical loss, which will be explained in Sec.~\ref{sec:align_loss} and \ref{sec:proto_loss}. }
\label{tabl:pipeline}
\end{table}

\subsection{Multimodal Network with a Transformer-based Fusion Module}
Our proposed multimodal network consists of two main components: unimodal backbones and a Transformer-based fusion module. The unimodal backbones consist of three separate feature extractors, which extract features from different input modalities. The fusion module aims to fuse and aggregate the features of different modalities from unimodal backbones and output the fused feature. There are two widely-used options for fusing modalities: using an MLP to process the concatenated representations of different modalities~\cite{poria2016convolutional, ortega2019multimodal, pandeya2021deep}, or utilizing a Transformer-based fusion module to take a series of tokens from different modalities~\cite{missingmodality,nagrani2021attention, siebert2022multi}. We adopt the Transformer-based fusion design as it can easily scale to an arbitrary number of input tokens using attention modules. This is especially important as the multimodal model is expected to handle data with a varying number of modalities in the context of our proposed task. The final output of the fusion module is obtained by averaging the output tokens instead of using the CLS token~\cite{vit,missingmodality}. Formally, the output of the fusion module $\boldsymbol{z}_{\text{fuse}}$ can be written as follows:
\begin{equation}
    \boldsymbol{z}_{\text{fuse}} = f\left(\left[\boldsymbol{x}_{\text{output}}^{m} + \boldsymbol{e}^m;| m \in \{\text{audio, video, IMU}\}\right]\right),
\end{equation}
where $\boldsymbol{x}_{\text{output}}^{m}$ represents the output representation of the feature extractor for modality $m$. $f$ is the fusion module that takes a sequence of input tokens from different modalities. Tokens from each modality are augmented with a modality-specific learnable embedding $\boldsymbol{e}^m$, which is used to disambiguate input tokens' modality information. 

During the training of the fusion module, we applied a technique named \emph{modality drop}. A subset of modalities is randomly dropped out with a probability $p$ during training, to ensure the robustness of the fusion module to a varying number of input modalities.

\subsection{Cross-Modal Alignment Multimodal Training}
\label{sec:align_loss}
In the zero-shot cross-modal setting, the multimodal model is required to learn and infer from disjoint modalities. One approach to achieving this is to construct a unified feature space that captures representations from different modalities. The feature space should ensure that features from the same data point but different modalities are in close proximity to each other. This allows knowledge learned from one modality to be applied to inference in other modalities. To achieve this, we propose to align features from the same data point but different modalities in multimodal training with contrastive loss.
Specifically, the unimodal feature output by the fusion module is represented as follows:
\begin{equation}
    \label{eq:out}
    \boldsymbol{z}_{m} = f\left(\boldsymbol{x}_{\text{output}}^{m}+ \boldsymbol{e}^m\right), \quad {m \in \{\text{audio, video, IMU}\}},
\end{equation}
which is expected to lie in the unified feature space.
We impose Noise Contrastive Estimation (NCE)~\cite{clip} loss to align video-audio and video-IMU pairs, drawn from different time stamps of video-audio-IMU data. Positive pairs consist of different modalities pairs from the same temporal location, while negative pairs are from different temporal locations. Our NCE alignment loss $\mathcal{L}_{\text{align}}$ is written as follows:
\begin{equation}
\begin{split}
&\mathcal{L}_{\text{align}}\left(\boldsymbol{z}_{\text{video}}, \boldsymbol{z}_{m}\right)= \sum_{m \in \{\text{audio, IMU}\}}\\
&-\log \left(\frac{\exp \left(\boldsymbol{z}_{\text{video}}^{\top} \boldsymbol{z}_{m} / \tau\right)}{\exp \left(\boldsymbol{z}_{\text{video}}^{\top} \boldsymbol{z}_{m} / \tau\right)+\sum_{z^{\prime} \in \mathcal{N}} \exp \left(\boldsymbol{z}_{\text{video}}^{\top} \boldsymbol{z}_{m}^{\prime} / \tau\right)}\right),
\end{split}
\end{equation}
where $\mathcal{N}$ are negative pairs in a batch. We use cosine similarity as the feature distance measurement metric in our NCS loss. $\tau$ is a temperature parameter controlling the softness. Unlike previous methods that build a hierarchical common space~\cite{akbari2021vatt}, our approach defines a unified feature space for all modalities.

\subsection{Cross-Modal Prototypical Loss}
\label{sec:proto_loss}
\emph{What properties can help representations better generalize in the few-shot task?} We design a novel extension of prototypical loss~\cite{prototypical}  that takes into account the alignment between features of different modalities.

\begin{figure}[!t]
    \centering
    \includegraphics[width=0.49\textwidth]{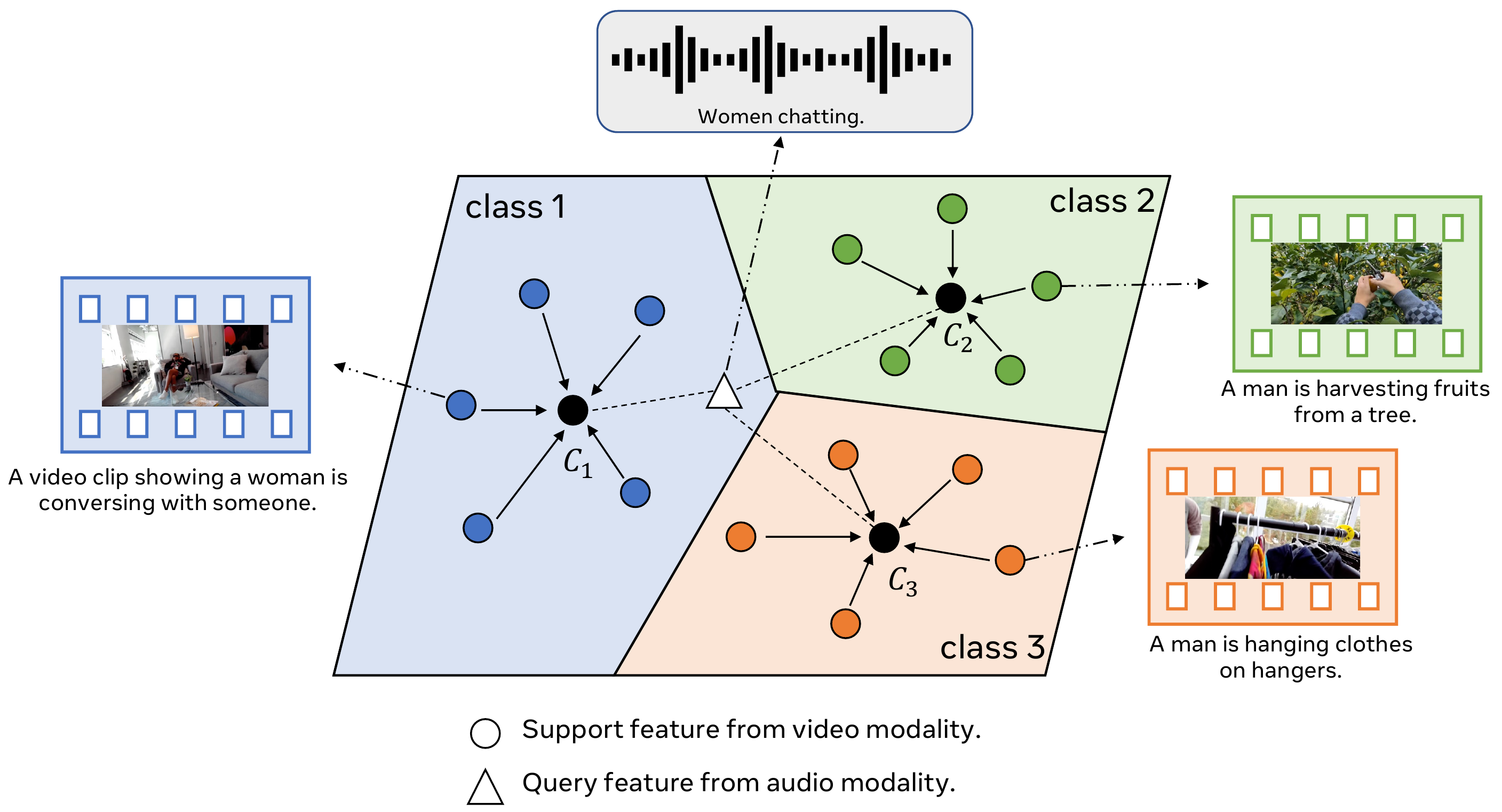}
       \caption{\textbf{Cross-modal prototypical loss}. Few-shot prototypes centroid $C_k$ computed by averaging support examples' feature. In contrast to the vanilla prototypical loss, our approach allows support and query examples to belong to different modalities. The figure shows an example where the support examples are video data, and the query example is audio data.}
    \label{fig:proto}
\end{figure}

The prototypical loss aims to minimize the distance between the centroid of support embeddings and the query embeddings in the feature space, where the labels for query data points are assigned according to their distance to every support centroid. In our proposed approach, support and query examples can belong to different modalities, allowing for cross-modal alignment (see Fig.~\ref{fig:proto}).
We use $\boldsymbol{z}^k_m$ to denote a unified space support feature of class $k$ and $\hat{\boldsymbol{z}}_n$  to represent a unified space query feature, where they might belong to different modalities $m, n \in \{\text{audio, video, IMU}\}$.
The centroid unified space support feature $c^k_m$ is calculated by averaging:
\begin{equation}
    c^k_m = \frac{1}{|\mathcal{Z}^k_m|} \sum_{\boldsymbol{z}^k_m \in \mathcal{Z}^k_m} \boldsymbol{z}^k_m,
\end{equation}
where $\mathcal{Z}^k_m$ is the set of support features of class $k$ with modality $m$.

The predicted probability of a query example $\hat{\boldsymbol{z}}_n$ belonging to class $k$ is computed using the negative exponential of the $\ell_2$ distance $d$ between the query feature and the centroid of the unified space support feature for class $k$:
\begin{equation}
P_k =\frac{\exp \left(-d\left(\hat{\boldsymbol{z}}_n, c^k_m\right)\right)}{\sum_{k^{\prime}} \exp \left(-d\left(\hat{\boldsymbol{z}}_n, c^{k^{\prime}}_m\right)\right)}, m, n \in \{\text{audio, video, IMU}\}
\end{equation}

Our proposed cross-modal prototypical loss $\mathcal{L}_{\text{proto}}$ is then formulated as the negative log-likelihood loss between the predicted probability and the ground truth class for the query example $\hat{y}$:
\begin{equation}
    \mathcal{L}_{\text{proto}} = \text{NLL}\left(\log\left[P_0, P_1, ... P_{N-1}\right], \hat{y}\right).
\end{equation}

In summary, our cross-modal prototypical loss extends the prototypical loss by enabling the cross-modal alignment between support and query features in the unified feature space. This loss can improve the generalization ability of representations in the zero-shot cross-modal task under the few-shot setting.

\vspace{-3pt}
\section{Experimental Setup}
\label{sec:experimental_setup}
\vspace{-3pt}
\subsection{Architecture Details}
\vspace{-3pt}
\noindent \textbf{Unimodal backbones}. We use MViT-B ($16 \times 4$) ~\cite{mvit} as the feature extractor for video modality, which is pre-trained on Kinetics-400~\cite{kinetics}. Audio Spectrogram Transformer (AST)~\cite{ast} is used as the audio feature extractor, and it is pre-trained on AudioSet~\cite{audioset}. For IMU feature extractor, we designed a ViT~\cite{vit} based transformer network. 

\noindent \textbf{Fusion module}.
Our fusion module is a transformer network with two layers. Each layer contains a self-attention block with 12 heads. The embedding dimension is $768$.

\subsection{Training \& Evaluation Details}
\vspace{-4pt}
We illustrate some basic details of the model training and evaluation. Hyper-parameters like learning rate and batch size are detailed in our supplementary material.

\noindent \textbf{Supervised setting}. Our model uses \emph{MMG-Ego4D} base classes for multimodal supervised training. Under the zero-shot cross-modal setting, our model also utilizes \emph{MMG-Ego4D} unlabeled data to do the multimodal unsupervised pre-training. We use Top-1 Accuracy to measure model performance.

\noindent \textbf{Few-Shot setting}. We use the finetune-based method to perform few-shot evaluation, where a small neural network is trained on the support set and is used to classify data points in the query set~\cite{guo2020broader, li2021improving, hu2022pushing}. We adopt the standard N-way K-shot setting~\cite{maml,vinyals2016matching} as the evaluation setting. Top-1 Accuracy is used to measure model performance. The final number is obtained by averaging the results on \num{10000} episodes. 
\vspace{-4pt}
\section{Results on \emph{MMG-Ego4D} Benchmark}
\label{sec:results}

\subsection{\emph{MMG-Ego4D} Few-Shot Setting Results}

%%%%%%%%%%%%%%%%%%%%%%%%%%
\begin{table}
\centering 
\resizebox{1.0\linewidth}{!}{
\begin{tabular}{c|cc|c|cc}
\toprule
\centering 
     \bf \multirow{2}{*}{Model} & \bf \multirow{2}{*}{FLOPs (G)}  & \bf \multirow{2}{*}{Param (M)}  & \bf \multirow{2}{*}{Modality}  & \bf \multirow{2}{*}{\shortstack[c]{5 Way 5 Shot\\ Accuracy}} & \bf \multirow{2}{*}{\shortstack[c]{Top-1 \\ Accuracy}}  \\
     & &  &  & &  \\
     
    \midrule
    MViT-B~\cite{mvit} & 70.50 & 36.50 & video & 58.89 & 52.40 \\
    AST~\cite{ast} & 42.08 & 87.03 & audio & 31.06 & 39.48 \\
    IMU Transformer & 1.65 & 15.55 & IMU & 40.07 & 29.78 \\ 

    \midrule 
    \bottomrule
\end{tabular}
}
\caption{\textbf{Unimodal few-shot \& supervised evaluation results}. Networks are trained on each modality independently. Video achieves the best performance, while also consuming more computational resources.}
\label{tabl:uni_results}
\end{table}

\begin{table}
\centering 
\resizebox{1.0\linewidth}{!}{
\begin{tabular}{c|ccc|ccc|c}
\toprule
\centering 
     \multirow{2}{*}{\shortstack[c]{\bf Eval.\\ \bf Setting}}  & \multicolumn{3}{c|}{\bf Support Modalities} &  \multicolumn{3}{c|}{\bf Query Modalities} & \multirow{2}{*}{\shortstack[c]{\bf 5 Way 5 Shot\\ \bf Accuracy}} \\

    \cmidrule(lr){2-4}
    \cmidrule(lr){5-7}
    % \cmidrule(lr){8}
     & Video & Audio & IMU & Video & Audio & IMU &  \\ %[0.2cm]
    \midrule

    Regular & \checkmark & \checkmark & \checkmark & \checkmark & \checkmark & \checkmark & 63.00  \\

    \midrule 
    
    \multirow{5}{*}{\rotatebox[origin=c]{90}{\thead{Missing \\ Modality}}} 
    & \checkmark & \checkmark & \checkmark & \checkmark & \checkmark &  & 61.76  \\
    & \checkmark & \checkmark & \checkmark &  & \checkmark & \checkmark  & 50.77  \\
    & \checkmark & \checkmark & \checkmark & \checkmark &  & \checkmark & 62.79  \\
    & \checkmark & \checkmark & \checkmark & \checkmark &  &  & 62.68 \\
    & \checkmark & \checkmark & \checkmark &  & \checkmark &  & 43.65 \\
    & \checkmark & \checkmark & \checkmark &  &  & \checkmark & 47.48 \\
    \midrule 
    
    \multirow{6}{*}{\rotatebox[origin=c]{90}{\thead{Zero-Shot \\ Evaluation}}} 
    &  & \checkmark &  & \checkmark &  &  & 46.90  \\
    &  &  & \checkmark & \checkmark &  &  & 42.07  \\
    &   & \checkmark & \checkmark & \checkmark &  &  & 50.80 \\
    & \checkmark &  &  &  & \checkmark & & 44.01 \\
    & \checkmark &  &  &  &  & \checkmark & 46.56  \\
    & \checkmark &  &  &  & \checkmark & \checkmark & 49.37 \\
    \midrule 
    \bottomrule
\end{tabular}
}
\caption{\textbf{Multimodal few-shot evaluation results}. These results are obtained with a single network that works across all three evaluation settings. We show the \textit{regular evaluation} results in the first block, where the model is trained and evaluated with all the modalities. The second block presents \textit{missing-modality results}, where the model is trained on all modalities but evaluated only on a subset. The last block is the result of \textit{cross-modal zero-shot} evaluation, where the training and evaluation modalities are disjoint. Note that all results are obtained using the same model weight. Our supplementary material provides results with more training and test modalities configurations.} 
\label{tabl:results_fewshot}
\end{table}
%%%%%%%%%%%%%%%%%%%%%%%%

\noindent \textbf{Multimodal system outperforms unimodal system significantly.} Tab.~\ref{tabl:uni_results} presents the few-shot classification results for individual modalities. Notably, the video modality achieves the highest accuracy, which is anticipated since most classes can be easily recognized using visual information. However, as illustrated in Fig~\ref{fig:multimodal}, fusing information from different modalities is critical to achieving better performance in egocentric action detection. Our proposed multimodal system outperforms the best-performing unimodal system by $4.11$ in terms of accuracy (Tab.~\ref{tabl:results_fewshot} block 1).

\noindent \textbf{Missing modality generalization}.
We present the results of the missing modality evaluation in the second block of Tab.~\ref{tabl:results_fewshot}, where the query modality is a subset of the support modality. Our model exhibits good generalizability even when some modalities are missing during evaluation, achieving solid accuracy. Notably, including the video modality in the query set yields a slight change in performance compared to the multimodal case. When video modality is not included, there is a $19.41\%$ drop in accuracy, indicating that video modality is the most informative. 
Surprisingly, when queries have only one cheap modality (audio or IMU), our method outperforms unimodal results (Tab. \ref{tabl:uni_results}) by a large margin of $19.24\%$ on IMU and $40.53\%$ on audio modality, demonstrating the effectiveness of our approach.

\noindent \textbf{Zero-shot cross-modal generalization}. This task presents a more significant challenge than missing modality generalization as the support and query modalities are disjoint. We select a few combinations and present the results in the last block of Tab.~\ref{tabl:results_fewshot}. To enable efficient training, we choose a setting where the support modality is computationally cheap, such as IMU and Audio, while the query modality is relatively more informative, such as video, to achieve high performance. Our model significantly outperforms the audio and IMU unimodal settings using this evaluation setting. We also present the results of using video as the support modality and IMU and/or audio as the query modality, where our model still obtains decent accuracy. While we did not include all support-query modality combinations in the paper due to space limitations, readers can refer to our supplementary materials for additional results.

\vspace{-2pt}
\subsection{\emph{MMG-Ego4D} Supervised Setting Results}
\vspace{-2pt}
The results of the supervised settings are presented in Tab. \ref{tabl:results_supervised}. Our multimodal model outperforms each unimodal model in Tab. \ref{tabl:uni_results} significantly in the regular setting.
Regarding the missing modality evaluation, our method exhibits strong generalization ability in the presence of missing modalities. If the video modality is preserved in the evaluation modality, the performance only experiences a minor drop. However, when video data is missing during evaluation, the performance drops by around $33\%$, suggesting that the video modality is more informative than the other two modalities.
The last block of Tab. \ref{tabl:results_supervised} shows the zero-shot cross-modal results. We explore two cases: using expensive modalities for training and cheap modalities for inference, and using cheap modalities for training and expensive modalities for inference. We observe that the model performs better in the latter case, indicating that learning from informative modalities benefits the model more.
\begin{table}
\centering 
\resizebox{1.0\linewidth}{!}{
\begin{tabular}{c|ccc|ccc|c}
\toprule
\centering 
     \multirow{2}{*}{\shortstack[c]{\bf Eval.\\ \bf Setting}}  & \multicolumn{3}{c|}{\bf Train Modalities} &  \multicolumn{3}{c|}{\bf Test Modalities} & \multirow{2}{*}{\shortstack[c]{\bf Top-1 \\ \bf Accuracy}} \\

    \cmidrule(lr){2-4}
    \cmidrule(lr){5-7}
     & Video & Audio & IMU & Video & Audio & IMU & \\ %[0.2cm]
    \midrule

    Regular & \checkmark & \checkmark & \checkmark & \checkmark & \checkmark & \checkmark  & 55.66  \\ %*

    \midrule 
    
    \multirow{3}{*}{\rotatebox[origin=c]{90}{\thead{Missing \\ Modality}}} 
    & \checkmark & \checkmark & \checkmark & \checkmark & \checkmark &   & 55.47  \\
    & \checkmark & \checkmark & \checkmark &  & \checkmark & \checkmark  & 37.07  \\
    & \checkmark & \checkmark & \checkmark & \checkmark &  & \checkmark  & 54.57  \\
    \midrule 
    
    \multirow{6}{*}{\rotatebox[origin=c]{90}{\thead{Zero-Shot \\ Evaluation}}} &  & \checkmark &  & \checkmark &  &   & 30.98\\
    &  &  & \checkmark & \checkmark &  &   & 20.00  \\
    &  & \checkmark & \checkmark & \checkmark &  &  & 25.03 \\
    & \checkmark  &  &  &  & \checkmark &  & 43.43 \\
    & \checkmark  &  &  &  &  & \checkmark & 35.67 \\
    & \checkmark  &  &  &  & \checkmark & \checkmark & 41.02 \\
    \midrule 
    \bottomrule
\end{tabular}
}
\caption{\textbf{Supervised setting evaluation results}. Results are organized following the same structure as in Tab. \ref{tabl:results_fewshot}. The model has the same weight in regular and missing modality evaluation.}
\label{tabl:results_supervised}
\end{table}

\subsection{Insights from Ablation Study}
\label{sec:ablation}
\begin{table*}[!ht]
\centering 
\resizebox{\linewidth}{!}{
\begin{tabular}{c|ccc|ccc|cc|c|c|c}
\toprule
\centering 
     \multirow{2}{*}{\shortstack[c]{\bf Eval.\\ \bf Setting}} & \multicolumn{3}{c|}{\bf Train/Support Modal.} &  \multicolumn{3}{c|}{\bf Test/Query Modal.} & \multirow{2}{*}{\shortstack[c]{\bf Fusion\\ \bf Module}} & \multirow{2}{*}{\shortstack[c]{ \bf Contrastive \\ \bf Alignment}} & \multirow{2}{*}{\shortstack[c]{ \bf Top-1 \\ \bf Accuracy}} & \multirow{2}{*}{\shortstack[c]{ \bf Cross-Modal\\ \bf Proto. Loss}} & \multirow{2}{*}{\shortstack[c]{\bf 5 Way 5 Shot\\ \bf Accuracy}}  \\

    \cmidrule(lr){2-4}
    \cmidrule(lr){5-7}
     & Video & Audio & IMU & Video & Audio & IMU & & & & & \\ 
    \midrule

    \multirow{3}{*}{Regular} & \multirow{3}{*}{\checkmark} & \multirow{3}{*}{\checkmark} & \multirow{3}{*}{\checkmark} & \multirow{3}{*}{\checkmark} & \multirow{3}{*}{\checkmark} & \multirow{3}{*}{\checkmark} & Attention & \checkmark & \bf 55.66 & \checkmark & \bf 63.00 \\ 
     &  &  &  &  &  &  & Attention  & \textcolor{red}{$\times$} & 52.18 & \checkmark & 61.16 \\
     &  &  &  &  &  &  & \textcolor{red}{MLP}  & \checkmark & 52.79 & \checkmark & 58.67 \\ 
     &  &  &  &  &  &  & Attention & \checkmark & - & \textcolor{red}{$\times$} & 62.37 \\ 
      \hline
     
     \multirow{3}{*}{\shortstack[c]{Missing\\ Modality}} & \multirow{3}{*}{\checkmark} & \multirow{3}{*}{\checkmark} & \multirow{3}{*}{\checkmark} & & \multirow{3}{*}{\checkmark} & \multirow{3}{*}{\checkmark} & Attention & \checkmark  & \bf 37.07 & \checkmark & \bf 50.77  \\ 
     &  &  &  &  &  & & Attention & \textcolor{red}{$\times$}  & 21.32 & \checkmark & 40.87 \\ 
    &  &  &  & &  &  & \textcolor{red}{MLP} & \checkmark  & 32.89 & \checkmark & 49.00 \\
    &  &  &  &  &  &  & Attention & \checkmark & - & \textcolor{red}{$\times$} & 50.03 \\ 
     \hline
     
     \multirow{3}{*}{\shortstack[c]{Zero-shot\\ Cross-Modal}} &  & \multirow{3}{*}{\checkmark} & \multirow{3}{*}{\checkmark} & \multirow{3}{*}{\checkmark} &  & & Attention & \checkmark  & \textbf{25.03}$^*$ & \checkmark & \bf 51.40 \\ 
    &  &  &  & &  &  & Attention  & \textcolor{red}{$\times$} & 2.37 & \checkmark & 33.93 \\
    &  &  &  & &  &  & \textcolor{red}{MLP}   & \checkmark  &  24.54$^*$ & \checkmark & 51.08 \\
    &  &  &  &  &  &  & Attention & \checkmark & - & \textcolor{red}{$\times$} & 50.80 \\ 
    % \midrule 
    
    \bottomrule
\end{tabular}
}
\vspace{-1pt}
\caption{\textbf{Ablation study of each design component under supervised \& few-shot settings}. Our proposed components improve the performance under all evaluation settings. Note that cross-modal prototypical loss is only applied under the few-shot setting.
$^*$Different from other settings, the cross-modal contrastive alignment loss is applied at the unsupervised multimodal pre-training stage in the supervised zero-shot cross-modal setting.}
\label{tabl:ablation}
\end{table*}

In this section, we carefully ablate the effect of each component in our designed multimodal system under various evaluation settings, including the regular, missing modality, and cross-modal zero-shot evaluations. 

\noindent \textbf{Fusion module}. 
In this study, we propose the use of a Transformer-based fusion module as an alternative approach to integrating information from different modalities in a multimodal network. To evaluate its performance, we conduct a comparative analysis against an MLP-based fusion module that concatenates representations from diverse modalities and processes them using an MLP. To ensure a fair comparison, we maintain the dimensionality of input and output representations of both modules to be consistent, with a similar number of parameters. In situations where some modalities are not present in the input of the MLP-based fusion module, we replace their representations with zero vectors. The results of the ablation study presented in Tab.~\ref{tabl:ablation} demonstrate that the Transformer-based fusion module outperforms the MLP-based fusion module in both few-shot and supervised learning scenarios across all tasks. We also investigate three decision choices empirically. Specifically, we examine the efficacy of using the CLS token or averaging all output tokens for the final prediction. We find that averaging all output tokens produces better performance. Additionally, we evaluate the inclusion of modality-specific embeddings before fusion and find that it is effective in aiding the model's ability to differentiate between modalities. Finally, we experiment with various dropout rates ($p$) for modality dropout and find that consistent performance is obtained across a range of values ($0.3$ to $0.8$), with the best results achieved at $p=0.6$.

\noindent \textbf{Cross-modal contrastive alignment loss}. Our motivation for incorporating cross-modal alignment loss into our pipeline is rooted in the desire to enhance cross-modal zero-shot generalization performance. In Tab.~\ref{tabl:ablation}, the inclusion of this component resulted in a remarkable improvement of $22.66$ and $17.47$ in cross-modal zero-shot generalization performance in supervised and few-shot learning settings, respectively. Additionally, we observed that the incorporation of cross-modal alignment loss also yielded performance gains in regular and missing modality tasks. These results underscore the importance of cross-modal alignment in succeeding in the \emph{MMG-Ego4D} benchmark.

\noindent \textbf{Cross-Modal prototypical loss}. In our study, we proposed the incorporation of cross-modal prototypical loss as a means of enhancing few-shot performance in MMG tasks. Our experimental results, as demonstrated in Tab.~\ref{tabl:ablation}, reveal that this novel component contributes to performance improvements of $0.74$ and $0.6$ points in missing modality and zero-shot scenarios, respectively, while also yielding an enhancement of $0.63$ points in the regular modality complete evaluation setting. These findings attest to the efficacy of cross-modal prototypical loss as a valuable addition to the MMG task performance optimization strategy.

\section{Conclusions}
In this paper, we introduced the first comprehensive benchmark for multimodal generalization (MMG) and proposed three components to improve the generalization performance of models. Our benchmark, \emph{MMG-Ego4D}, includes two new tasks and a new dataset. The evaluation of different baseline architectures showed that the generalization ability of current systems is limited. Therefore, benchmarking and improving generalization ability deserve attention, especially as models are deployed into more sensitive use cases. Through extensive experiments and ablation study, we demonstrated that our proposed attention-based fusion mechanism with modality dropout training and alignment of unimodal representation during fusion could improve the performance of supervised and few-shot tasks in \emph{MMG-Ego4D}. Our proposed cross-modal prototypical loss also improves the performance of few-shot tasks in \emph{MMG-Ego4D}. We created a new dataset and introduced novel experiments for the rigorous study of multimodal generalization problems. These methods can increase generalizability and are essential for real-world settings where secure environemnts are important. 

\clearpage
%%%%%%%%% REFERENCES
{\small
\bibliographystyle{ieee_fullname}
\bibliography{arxiv}
}

\end{document}